\documentclass[journal]{IEEEtran}

\ifCLASSINFOpdf
\else
   \usepackage[dvips]{graphicx}
\fi
\usepackage{url}

\usepackage{graphicx}
\usepackage{booktabs}       
\usepackage{amsfonts}      
\usepackage{nicefrac}      
\usepackage{microtype}    
\usepackage{amsmath}
\usepackage{bbm}
\usepackage{amstext}
\usepackage{amssymb}
\usepackage{graphicx}
\usepackage{float}
\usepackage{subfigure}
\usepackage{placeins}
\usepackage{multicol}
\usepackage{algorithm}
\usepackage{algorithmic}
\usepackage{wrapfig}
\usepackage{multirow}
\usepackage[table]{xcolor}
\usepackage{bigstrut}
\usepackage[medium,compact]{titlesec}
\usepackage{hyperref}

\begin{document}
\title{Enhancing Adversarial Transferability via Information Bottleneck Constraints}

\author{
	Biqing Qi\textsuperscript{1,2,3}, Junqi Gao\textsuperscript{4}, Jianxing Liu\textsuperscript{1}, \emph{Senior Member}, \emph{IEEE}, Ligang Wu\textsuperscript{1}, \emph{Fellow}, \emph{IEEE}, Bowen Zhou\textsuperscript{1,}\textsuperscript{2} \emph{Fellow}, \emph{IEEE}
\thanks{
\textsuperscript{1}Department of Control Science and Engineering, Harbin Institute of Technology, Harbin 150001, P. R. China; \textsuperscript{2} Tshinghua University, Bejing, P. R. China; \textsuperscript{3}Frontis.AI, Beijing, P. R. China; \textsuperscript{4}School of Mathematics, Harbin Institute of Technology, Harbin 150001, P. R. China; ( Email: qibiqing7@gmail.com; gjunqi97@gmail.com; jx.liu@hit.edu.cn; ligangwu@hit.edu.cn; zhoubowen@tsinghua.edu.cn). 
This work was supported in part by the National Science and Technology Major Project (No. 20232D0121403).
\textit{Corresponding authors: Bowen Zhou and Ligang Wu.}

} 

}

\maketitle

\begin{abstract}
From the perspective of information bottleneck (IB) theory, we propose a novel framework for performing black-box transferable adversarial attacks named IBTA, which leverages advancements in invariant features. Intuitively,
diminishing the reliance of adversarial perturbations on the original data, under equivalent attack performance constraints, encourages a greater reliance on invariant features that contributes most to classification, thereby enhancing the transferability of adversarial attacks.
Building on this motivation, we redefine the optimization of transferable attacks using a novel theoretical framework that centers around IB.
Specifically, to overcome the challenge of unoptimizable mutual information, we propose a simple and efficient mutual information lower bound (MILB) for approximating computation.
Moreover, to quantitatively evaluate mutual information, we utilize the Mutual Information Neural Estimator (MINE) to perform a thorough analysis. Our experiments on the ImageNet dataset well demonstrate the efficiency and scalability of IBTA and derived MILB. Our code is available at
\href{https://github.com/Biqing-Qi/Enhancing-Adversarial-Transferability-via-Information-Bottleneck-Constraints}{github.com/IBTA.}
\end{abstract}


\IEEEpeerreviewmaketitle
\section{Introduction}
\IEEEPARstart{D}{eep} learning-based models are susceptible to imperceptible adversarial examples (AEs)
\cite{szegedy2013intriguing}, which often exhibit successful transfer across different models\cite{goodfellow2014explaining,liu2016delving,QiZZLW24,GaoQLGLXZ23}. This characteristic provides a practical advantage in black-box scenarios, where the parameters and structure of the victim model remain unknown.
Studying transferable attacks contributes to a deeper comprehension of adversarial vulnerabilities and can offer valuable insights for the development of defense strategies.
However, traditional black-box attacks demonstrate limited performance in terms of transferability \cite{liu2016delving,carlini2017towards,madry2018towards}.
Nowadays, various studies delved into transfer attacks and proposed strategies aimed at enhancing the transferability of adversarial attacks. The mainly works can be classified according to their objectives as either targeted or non-targeted attacks. Targeted attacks aim at a specific class, while non-targeted attacks aim to disrupt any class except the original one.
In the non-target scenario, numerous strategies involve the incorporation of momentum in the iterative process \cite{dong2018boosting,lin2019nesterov}, the utilization of data augmentation \cite{dong2019evading,zou2020improving}, and the application of mixup techniques \cite{wang2021admix}. 
Actually, transferring targeted adversarial perturbations poses greater challenges compared to non-targeted settings \cite{li2020towards,zhao2021success}, prompting subsequent research endeavors dedicated to investigating the transferability of targeted attacks.
These efforts encompass the use of GAN-generated adversarial perturbations \cite{naseer2021generating}, the harnessing of local and global information to enhance contrast  \cite{wei2023enhancing}, the incorporation of 3D transformations \cite{byun2022improving} into attack strategies, the utilization of spectral transformations, and the formulation of the iterative process as a min-max bi-level optimization problem.
Despite extensive research, non-targeted and targeted investigations remain separate. 
Therefore, a unified perspective for comprehending adversarial transferability is still absent.

To response the above challenge, we introduce a novel perspective for enhancing transferability in adversarial attacks, leveraging the principles of information bottleneck (IB) theory \cite{tishby2000information,tishby2015deep}. Our contributions are three-fold.
\textbf{Firstly}, 
We are the first to introduce a simple and efficient framework to enhance transferability within adversarial attacks from the information bottleneck perspective. 
This framework originates from the motivation to diminish correlation on noisy and extraneous features present in the original inputs, thereby improving the transferability of adversarial perturbations. 
It is formulated by mutual information (MI) techniques.
To the best of our knowledge, this conceptual perspective remains unexplored in previous studies.
\textbf{Secondly}, because direct optimization of the MI between perturbations and the original input is unfeasible, inspired from key ideas in variational inference \cite{alemi2016deep,wainwright2008graphical}, we derived a lower bound for this measure. Following that, we introduce an efficient loss function known as the IB based transferable loss to facilitate the suppression of MI.
\textbf{Thirdly}, to quantitatively assess the lower bound of MI, we employed the Mutual Neural Estimator (MINE) technique \cite{belghazi2018mutual}. We conducted experimental validation to ascertain the effectiveness of our method in suppressing MI. By conducting experiments on the ImageNet dataset, we illustrate that our method consistently enhances performance across various baselines.

\subsection{Motivation}

In the context of deep learning models, a fundamental question arises: how do we extract meaningful features from the original data? These features have a significant impact on the model's performance. Inspired by prior works \cite{tishby2000information,tishby2015deep}, we recognize that invariant features are closely aligned with class information, while non-invariant features consist of noise and instability.  
Intuitively, invariant features that capture essential information are more versatile, thus enhancing transferability.
This naturally leads us to inquire: \textbf{Can emphasizing the correlation between attacks and invariant features enhance adversarial transferability?}
This proposition suggests that if adversarial perturbations primarily depend on invariant features while reducing reliance on non-invariant features and noise in the original data, they will demonstrate superior transferability across a diverse range of models.
To achieve this, we apply the IB Theory, which entails a dual purpose: maximizing information extraction from the encoded hidden layer variable denoted as $Z$ with a specific emphasis on the included label $Y$ while ensuring task completion. Simultaneously, we aim to minimize the information related to the input variable $X$ within $Z$, which can be accomplished by minimizing $I(X; Z) - \beta I(Y; Z)$. 

\subsection{Problem Formulation}
Formally, in line with the aforementioned motivation, we proceed to reformulate adversarial attacks with the help of IB concepts.
With the scope of adversarial attacks, consider a clean sample $(\boldsymbol x,y)\in \Omega$, where $\Omega = \mathcal X\times \mathcal Y$ represents the sample space, and the source model $\mathcal M^{\mathcal S}(\cdot, {\boldsymbol {\theta}})$ is parameterized by ${\boldsymbol {\theta}}\in \Theta$. We can redefine an adversarial attack as follows:
\vspace{-2pt}
\begin{equation}
\label{eq1}
    \underset{\|\boldsymbol \epsilon\| \le r}{\text{minimize}} \quad \zeta_{tar} \mathcal L_{cls}(\mathcal M^{\mathcal S}(\boldsymbol x+\boldsymbol \epsilon, {\boldsymbol {\theta}}), y_{adv})+I(\mathcal E;X|Y_{adv}),
\end{equation}
\vspace{-2pt}
in which $r$ denotes the constraint threshold for the adversarial perturbation, and $\mathcal E$ represents the random variable associated with this perturbation. The variable $\zeta_{tar}$ takes the value $1$ in the target setting and $-1$ otherwise, while $\mathcal L_{cls}$ signifies the classification loss. $y_{adv}$ indicates the label employed during the attack. In the targeted scenario, $y_{adv}$ corresponds to the target class label $y_{target}$; otherwise, it defaults to the label $y$.

\subsection{IB induced Transferable Attacks (IBTA)}

Given the optimization objective in Eq. (1), direct optimization of $I(\mathcal E; X | Y_{adv})$ is infeasible. To address this challenge, we derive a simple lower-bound to realize approximate computation. 
Firstly, we consider the information flow during the attack process: $X\rightarrow \mathcal E \rightarrow X_{adv}\rightarrow Z_{adv}^{\boldsymbol \theta}$, suppose it forms a Markov chain. Since $\boldsymbol{z}_{adv}^{\boldsymbol \theta}=\mathcal M^{\mathcal S}(\boldsymbol{x}_{adv}, {\boldsymbol {\theta}})$ (Adversarial perturbation $\boldsymbol z_{adv}$ in the source side can be easily guaranteed), the information flow becomes $X \rightarrow \mathcal E \rightarrow Z_{adv}^{\boldsymbol \theta}$. Expanding MI in two distinct ways:
\vspace{-2pt}
\begin{align*}
&I(X;\mathcal E, Z_{adv}^{\boldsymbol \theta} | Y_{adv})=I(X; Z_{adv}^{\boldsymbol \theta} | Y_{adv}) +I(X; Z_{adv}^{\boldsymbol \theta} | Y_{adv}, \mathcal E)\\
&I(X;\mathcal E, Z_{adv}^{\boldsymbol \theta} | Y_{adv})=I(X; \mathcal E | Y_{adv}) +I(X; \mathcal E | Y_{adv}, Z_{adv}^{\boldsymbol \theta}).\tag{2}
\end{align*}

\begin{algorithm}
\caption{Procedure of IBTA}          
\begin{algorithmic}[1]
\REQUIRE Surrogate model $\mathcal M_{\mathcal S}(\cdot, \boldsymbol {\boldsymbol {\theta}})$, total steps $T$, step size $\alpha$, clean input $\boldsymbol x$, label used for attack $y_{adv}$, $\ell_\infty$ perturbation budget $r$.
\STATE Randomly sample $\boldsymbol \eta\leftarrow \mathcal N(0,\sigma^2\mathbf I)$, $\boldsymbol {\tilde x}\leftarrow \boldsymbol x+\boldsymbol \eta $
\STATE Initialize $\boldsymbol \epsilon_0 = 0$, $\boldsymbol{x}_{adv} \leftarrow \boldsymbol x+\boldsymbol \epsilon_0$, $\boldsymbol {\tilde x}_{adv} \leftarrow \Pi _{(0,1)}(\boldsymbol {\tilde x}+\boldsymbol \epsilon_0)$ 
\FOR{$t=1\leftarrow T$}
\STATE $g_t = \nabla_{\boldsymbol \epsilon} \mathcal L_{IBTA}(\boldsymbol{z}_{adv},\boldsymbol {\tilde z}_{adv}, y_{adv})$.
\STATE $\boldsymbol \epsilon_t = \Pi _{(-r,r)}(\boldsymbol \epsilon_{t-1} - \alpha*\text{sign}(g_t))$
\STATE $\boldsymbol{x}_{adv}=\Pi _{(0,1)}(x+\boldsymbol \epsilon_t), {\tilde x}_{adv}=\Pi _{(0,1)}({\tilde x}+\boldsymbol \epsilon_t)$
\ENDFOR
\RETURN $\boldsymbol{x}_{adv}$
\end{algorithmic}
\label{algorithm 1} 
\end{algorithm}
\vspace{-10pt}
Markovity implies that $X$ and $Z_{adv}^{\boldsymbol \theta}$ are conditionally independent
given $\mathcal E$, thus $I(X; Z_{adv}^{\boldsymbol \theta} | Y_{adv}, \mathcal E)=0$. Due to the non-negativity of MI, we have:
\vspace{-2pt}
\begin{equation}
\label{eq2}
I(X, \mathcal E | Y_{adv}) \ge I(X, Z_{adv}^{\boldsymbol \theta} | Y_{adv}). \tag{3}
\end{equation}
Note that:
\vspace{-2pt}
\begin{align*}
&I(X, Z_{adv}^{\boldsymbol \theta} | Y_{adv})\\
&=\int p(\boldsymbol{z}_{adv}^{\boldsymbol \theta},\boldsymbol x,y_{adv})\log\frac{p(\boldsymbol{z}_{adv}^{\boldsymbol \theta}|\boldsymbol x,y_{adv})}{p(\boldsymbol{z}_{adv}^{\boldsymbol \theta}|y_{adv})}d\boldsymbol{z}_{adv}^{\boldsymbol \theta}dy_{adv}d\boldsymbol x\\
&=\mathbb E_{X,Y_{adv}}\left[\mathcal D_{\mathbf{KL}}(p(\boldsymbol{z}_{adv}^{\boldsymbol \theta}|\boldsymbol x,y_{adv})\Vert p(\boldsymbol{z}_{adv}^{\boldsymbol \theta}|y_{adv})) \right]. \tag{4}
\end{align*}
\vspace{-2pt}
We can minimize the lower bound of $I(X, \mathcal E | Y_{adv})$ by minimizing $\mathcal D_{\mathbf{KL}}(p(\boldsymbol{z}_{adv}^{\boldsymbol \theta}|\boldsymbol x,y_{adv})\Vert p(\boldsymbol{z}_{adv}^{\boldsymbol \theta}|y_{adv}))$ for all $\boldsymbol x\in\mathcal X$, which means we aim to make $p(\boldsymbol{z}_{adv}^{\boldsymbol \theta}|\boldsymbol x,y_{adv})$ as close as possible to $p(\boldsymbol{z}_{adv}^{\boldsymbol \theta}|y_{adv})$ for all $\boldsymbol x\in\mathcal X$.
According to the definition of marginal distribution, we have 
\begin{align*}
p(\boldsymbol{z}_{adv}^{\boldsymbol \theta}|y_{adv})&=\int p(\boldsymbol{z}_{adv}^{\boldsymbol \theta},\boldsymbol x|y_{adv})dx\\
&=\int p(\boldsymbol{z}_{adv}^{\boldsymbol \theta}|\boldsymbol x, y_{adv})p(\boldsymbol x|y_{adv})d\boldsymbol x\\
&=\int p(\boldsymbol{z}_{adv}^{\boldsymbol \theta}|\boldsymbol x, y_{adv})p(\boldsymbol x)d\boldsymbol x\\
& = \mathbb E_X\left[p(\boldsymbol{z}_{adv}^{\boldsymbol \theta}|\boldsymbol x, y_{adv})\right].\tag{5}
\end{align*}
\vspace{-2pt}
Therefore, our goal becomes to make $p(\boldsymbol{z}_{adv}^{\boldsymbol \theta}|\boldsymbol x,y_{adv})$ as close as possible to $\mathbb E_X\left[p(\boldsymbol{z}_{adv}^{\boldsymbol \theta}|\boldsymbol x, y_{adv})\right]$. We can achieve this by reducing the variability of $p(\boldsymbol{z}_{adv}^{\boldsymbol \theta}|\boldsymbol x, y_{adv})$ with respect to $\boldsymbol x$. In other words, we need that $p(\boldsymbol{z}_{adv}^{\boldsymbol \theta}|\boldsymbol x, y_{adv})$ and $p(\boldsymbol {\tilde z}^{\boldsymbol \theta}_{adv}|\boldsymbol {\tilde x}, y_{adv})$ are as close as possible for different $\boldsymbol x, \boldsymbol {\tilde x} \in \mathcal X$.
To achieve this, we introduce a tiny displacement $\boldsymbol \eta$ to $\boldsymbol x$ to simulate $\boldsymbol {\tilde x}$, where $\boldsymbol \eta\sim\mathcal N(0,\sigma^2\mathbf I)$ and $\sigma$ is a hyperparameter with a small value, aiming to ensure that $\boldsymbol x+\boldsymbol \eta$ closely approximates the distribution $p(\boldsymbol x)$. Following this, we can introduce the IB induced Loss as depicted below, to help reduce $I(X, Z_{adv}^{\boldsymbol \theta} | Y_{adv})$:
\vspace{-2pt}
\begin{align*}
&\mathcal L_{IBTA}(\boldsymbol{z}^{\boldsymbol \theta}_{adv},\boldsymbol {\tilde z}^{\boldsymbol \theta}_{adv}, y_{adv})\\
&=\mathcal L(\boldsymbol{z}^{\boldsymbol \theta}_{adv},\boldsymbol {\tilde z}^{\boldsymbol \theta}_{adv}, y_{adv})+\gamma \mathcal L(\boldsymbol {\tilde z}^{\boldsymbol \theta}_{adv},\boldsymbol{z}^{\boldsymbol \theta}_{adv}, y_{adv}), \tag{6}
\end{align*}
\vspace{-2pt}
and
\vspace{-2pt}
\begin{align*}
&\mathcal L(\boldsymbol{z}^{\boldsymbol\theta}_{adv},\boldsymbol {\tilde z}^{\boldsymbol\theta}_{adv}, y_{adv})=\delta_{tar}\mathcal L_{cls}(\boldsymbol{z}^{\boldsymbol\theta}_{adv},y_{adv})\\
&+\lambda[\mathcal D_{\mathbf{KL}}(p(\boldsymbol{z}_{adv}^{\boldsymbol \theta}|\boldsymbol x,y_{adv})\Vert p(\boldsymbol {\tilde z}^{\boldsymbol \theta}_{adv}|\boldsymbol {\tilde x},y_{adv})), \tag{7}
\end{align*}
\vspace{-2pt}
where $\gamma$ and $\lambda$ are hyperparameters and blue{$\boldsymbol{z}_{adv}^{\boldsymbol \theta}=\mathcal M_{\mathcal S}(\boldsymbol x+\boldsymbol \epsilon, \boldsymbol {\boldsymbol {\theta}})$ and $\boldsymbol {\tilde z}^{\boldsymbol \theta}_{adv}=\mathcal M_{\mathcal S}(\boldsymbol {\tilde x}+\boldsymbol \epsilon, \boldsymbol {\boldsymbol {\theta}})$}, the classification loss $\mathcal L_{cls}$ is used to iteratively compute $\boldsymbol \epsilon$. The KL divergence is employed to ensure the proximity between $p(\boldsymbol{z}_{adv}^{\boldsymbol \theta}|\boldsymbol x, y_{adv})$ and $p(\boldsymbol {\tilde z}^{\boldsymbol \theta}_{adv}|\boldsymbol {\tilde x}, y_{adv})$. 
Based on this, we introduce the \textbf{IB induced Transferable Attack} (IBTA), its framework is summarized as Algorithm 1. The notation $\Pi _{(a,b)}(\cdot)$ denotes the operation of clipping each element's value to fall within the range $(a,b)$.
\vspace{-2pt}

\section{Analysis and Experiments}
\subsection{Analysis of MI}
\label{subsecC1}
 \textbf{Impact of $I(\mathcal E;X|Y_{adv}).$} To quantitatively evaluate the efficiently of our framework for reducing $I(\mathcal E;X|Y_{adv})$, we employed the MINE technique \cite{belghazi2018mutual} with the help of an $M$-bounded estimator $T_\varphi$ parameterized by $\varphi$ to compute $I_\varphi(\mathcal E;X|Y_{adv})$, i.e. $e^{T_\varphi},T_\varphi\le M$. $\varphi\in \Phi\subset \mathbb R^d $, $\|\varphi\|\le K$, and $T_\varphi$ is $L$-Lipschitz. As the training sample size $n$ increases, the estimated $I_\varphi(\mathcal E;X|Y_{adv})$ will method the true $I(\mathcal E;X|Y_{adv})$ from below, which is ensured by Theorem 1.
\newtheorem{theorem}{Theorem}
\vspace{-2pt}
\begin{theorem}(Theorem 3 in \cite{belghazi2018mutual})
\label{theorem.1}
Given any values $\xi$, $\delta$ of the desired accuracy
and confidence parameters, we have
\begin{align*}
\left.\operatorname{Pr}(\mid \widehat{I(X ; Z)})_{n}-I_{\Theta}(X, Z) \mid \leq \xi\right) \geq 1-\delta,  \tag{8}
\end{align*}
whenever the number $n$ of samples satisfies
\begin{align*}
n \geq \frac{2 M^{2}(d \log (16 K L \sqrt{d} / \xi)+2 d M+\log (2 / \delta))}{\xi^{2}}. \tag{9}
\end{align*}

\end{theorem}
\vspace{-2pt}

 \vspace{-5pt}

In our experiments, we conducted a random selection of ten classes (31, 56, 241, 335, 458, 532, 712, 766, 887, 975) from the ImageNet dataset \cite{deng2009imagenet} for training purposes. The estimator was trained using the respective training dataset, and adversarial perturbations were computed for this training data. Experiments were performed in both targeted and non-targeted settings using three classical baseline attack methods: MIM \cite{dong2018boosting}, DIM \cite{zou2020improving}, and TIM \cite{dong2019evading}. The values of $\lambda$, $\gamma$, and $\sigma$ were set to $0.01$, $0.1$, and $0.1$ respectively. We employed ResNet50 (Res50) \cite{he2016deep} as our source model. The perturbation budget was defined as $r = 16$, with $\alpha = 2$ and $T = 20$. The experimental results, as depicted in Figure 2, illustrate the effectiveness of our method in reducing $I(\mathcal E;X|Y_{adv})$.

\vspace{-2pt}
\begin{figure}[t]
\centering  
\subfigure{
\label{Fig1.sub.1}
\includegraphics[width=0.18\textwidth]{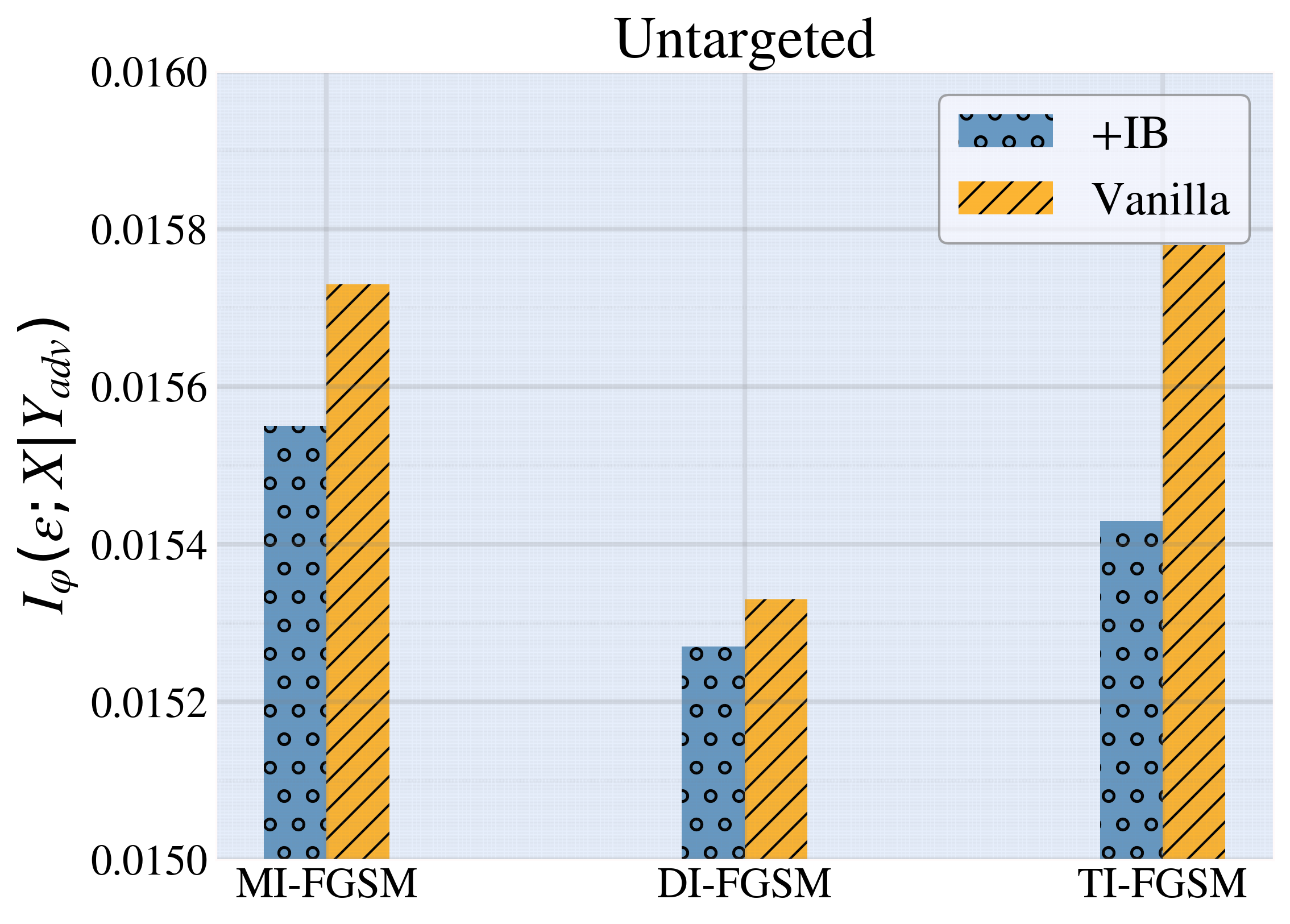}}
\subfigure{
\label{Fig1.sub.2}
\includegraphics[width=0.18\textwidth]{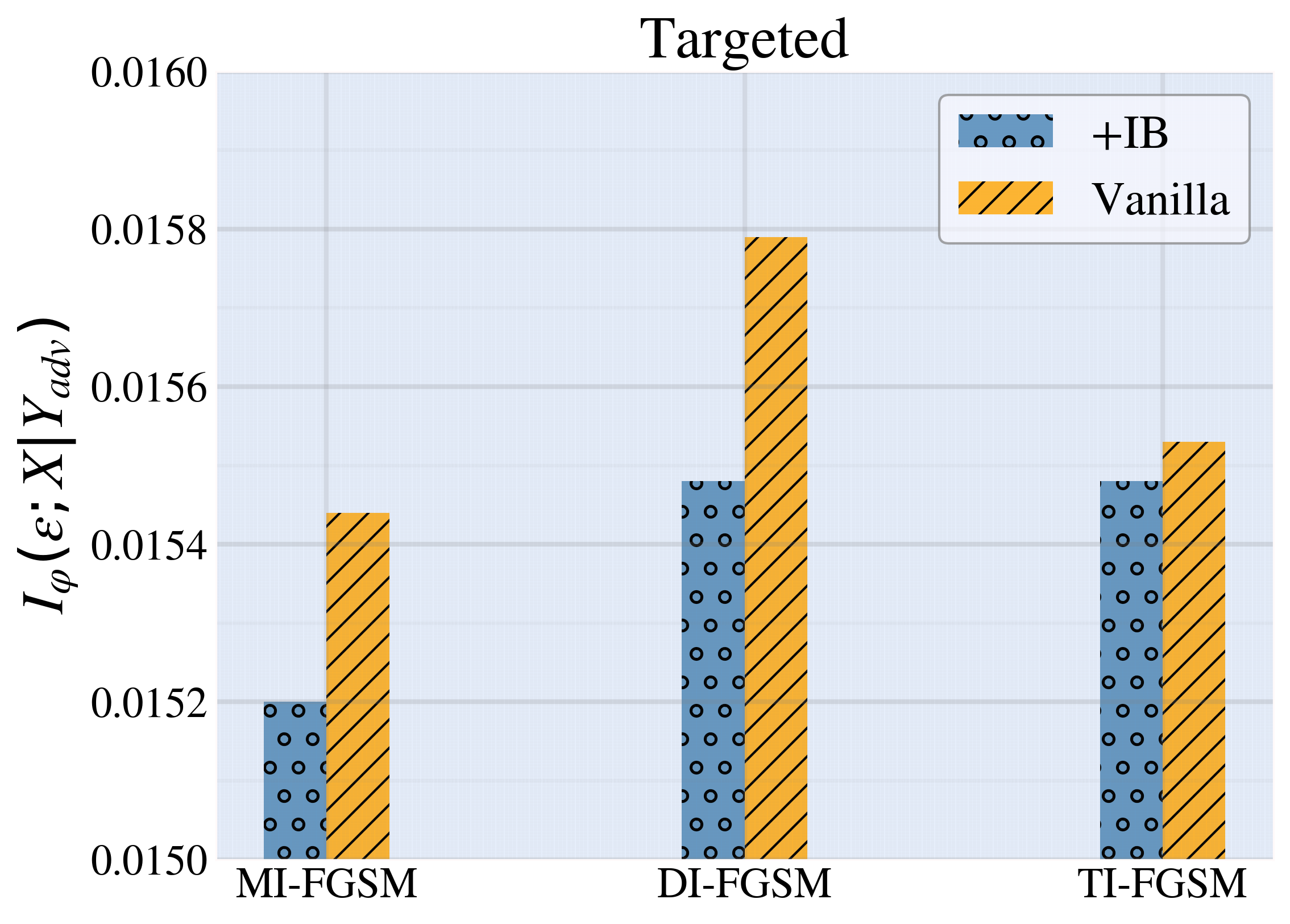}}

\vspace{-5pt}
\caption{The values of computed $I_\varphi(\mathcal E;X|Y_{adv})$ under targeted and non-targeted settings, with "+IBTA" indicating the addition of IBTA during the iteration process.}
\label{Fig 1}
\vspace{-10pt}
\end{figure}

\textbf{Impact of hyperparameters.} To investigate the impact of various hyperparameter choices on our method, we conducted comparative experiments using the MIM+IBTA method with different hyperparameter magnitudes. The source model employed was ResNet50 (Res50), while the target models evaluated included ResNet152 (Res152) \cite{he2016deep}, DenseNet121 (Dense121) \cite{huang2017densely}, and VGG19bn \cite{simonyan2014very}. We calculated the average transfer success rate across these three models. All other experimental settings remained consistent with those detailed in Section III.A. The results are presented in Figures 3 and 4.
When $\sigma$ was set to $1$, a decrease in transferability was observed in both target and non-targeted scenarios, with a more significant reduction in the non-targeted scenario. This phenomenon stems from the substantial deviation of $\boldsymbol x + \boldsymbol \eta$ from the original distribution $p(\boldsymbol x)$ when sigma is excessively large.
In non-targeted scenarios, choosing a larger $\lambda$ often resulted in a lower transfer success rate. This is because, in targeted scenarios, it suffices to include as much information about the target class as possible in $\boldsymbol \epsilon$. However, in non-targeted scenarios, we need to perturb $\boldsymbol x$ and $\boldsymbol {\tilde x}$ in directions that maximize their original class classification losses, which are likely to be inconsistent. A larger lambda exacerbates this imbalance.
In non-targeted settings with $\sigma=1$, increasing gamma led to some performance degradation, whereas in targeted settings with $\sigma=0.01$, it yielded a certain degree of improvement. For stability, we opted for $\lambda=0.1$, $\sigma=0.1$, and $\gamma=0.1$ in subsequent experiments.

 \vspace{-3pt}
\begin{figure}[t]
\centering  
\subfigure{
\label{Fig2.sub.1}
\includegraphics[width=0.15\textwidth]{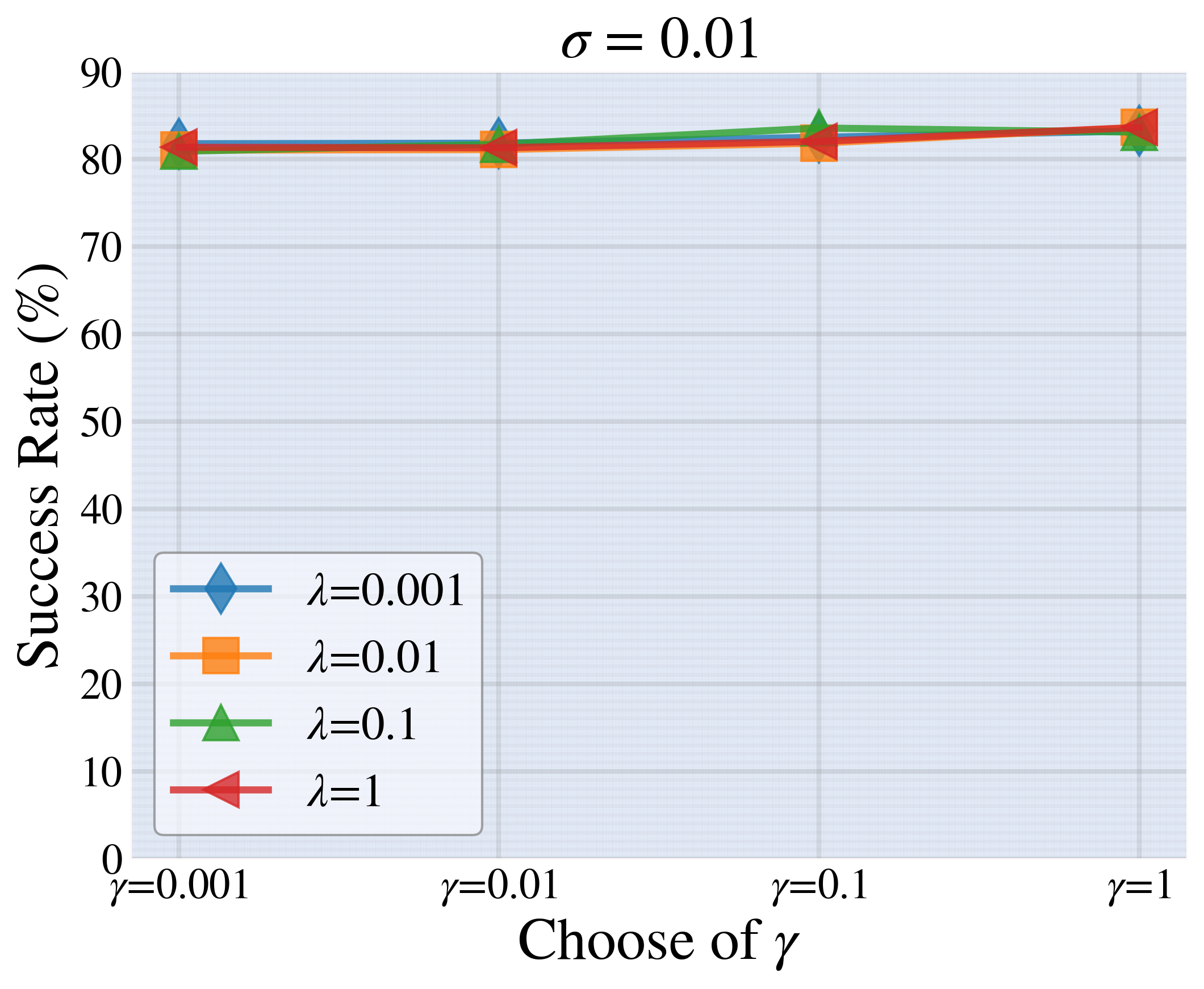}}
\subfigure{
\label{Fig2.sub.2}
\includegraphics[width=0.15\textwidth]{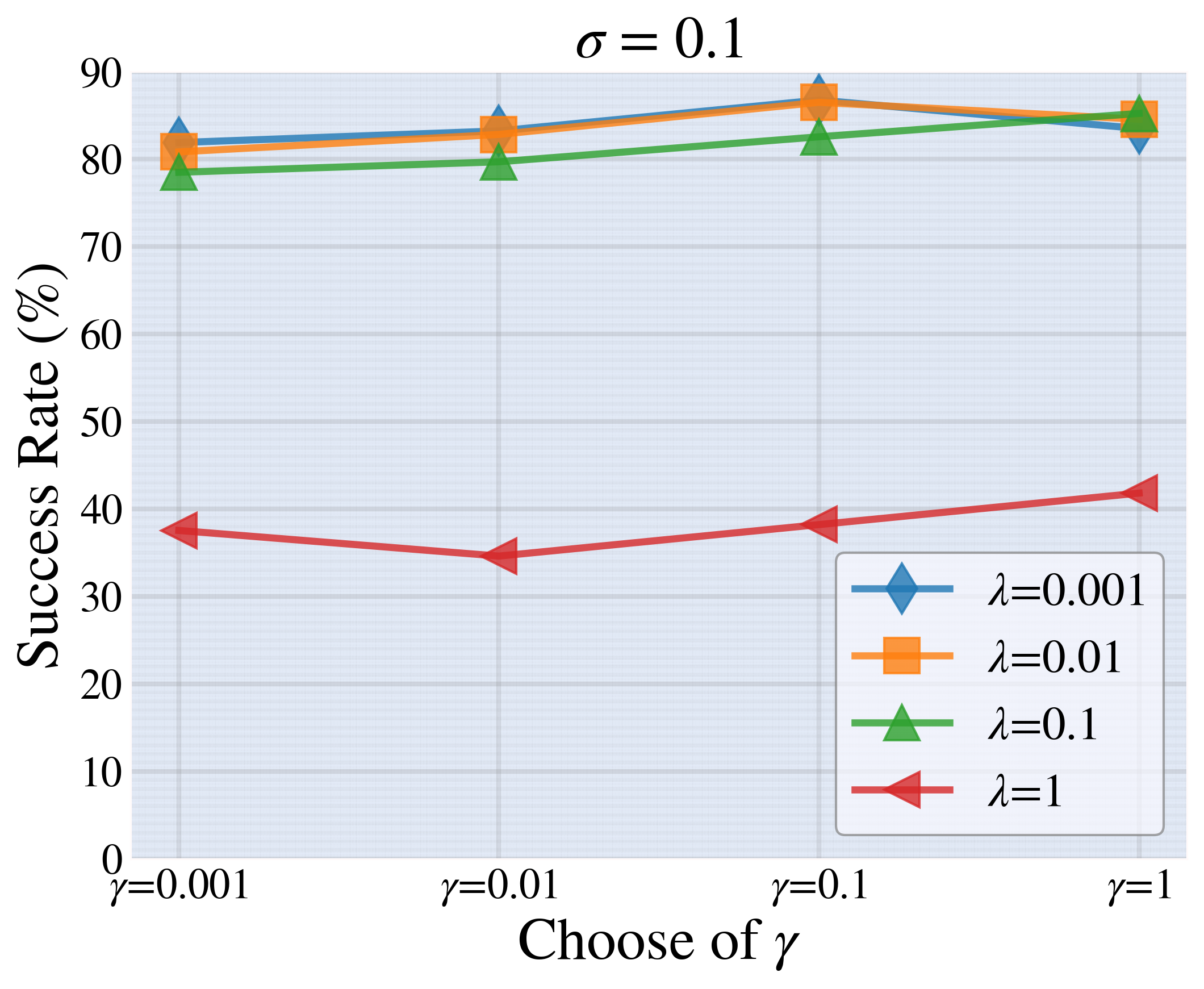}}
\subfigure{
\label{Fig2.sub.3}
\includegraphics[width=0.15\textwidth]{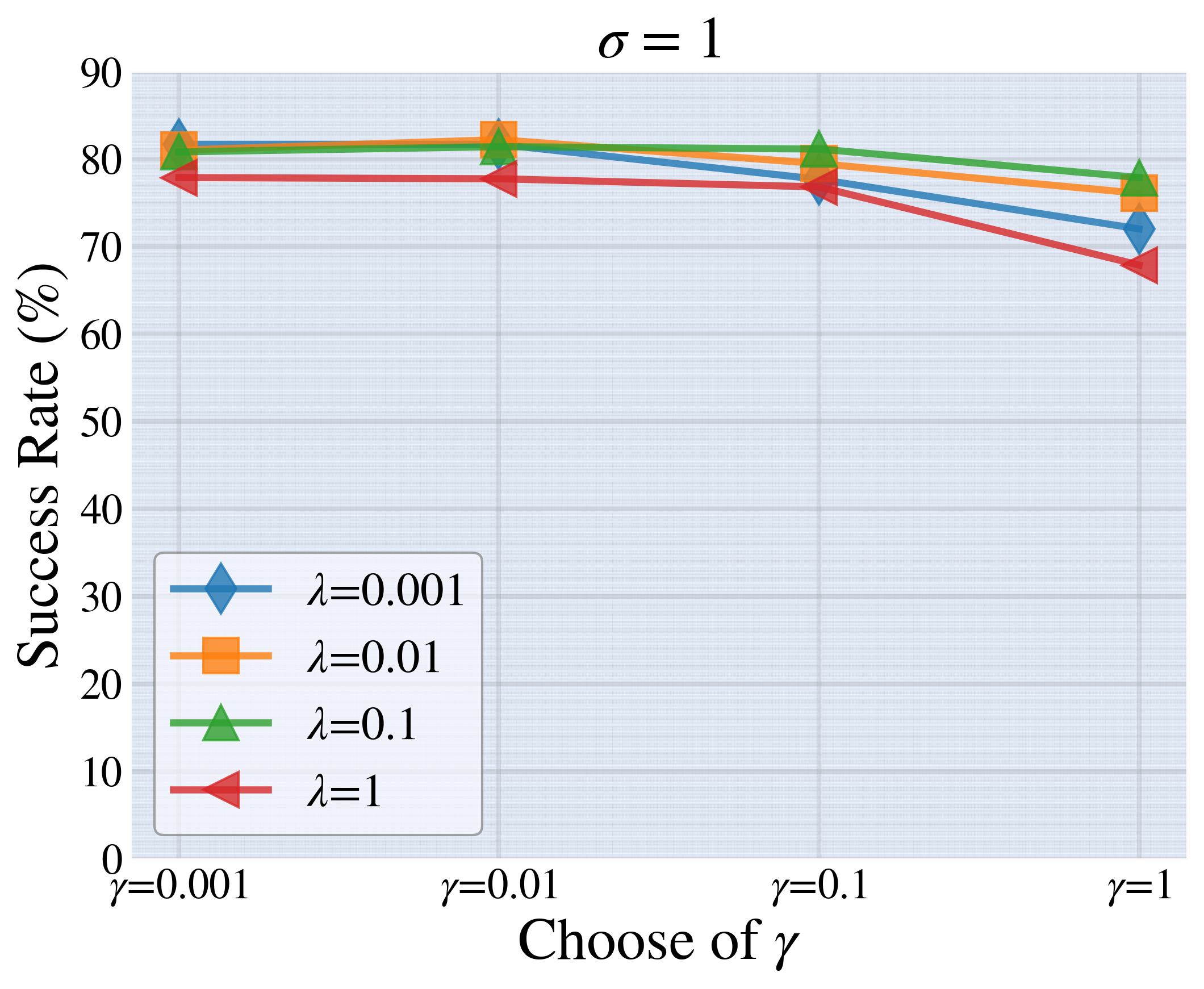}}
\vspace{-5pt}
\caption{Transfer success rates of MIM+IBTA with different parameter settings in the untargeted scenario.}
\label{Fig 2}
\vspace{-5pt}
\end{figure}

\vspace{-3pt}

\begin{figure}[t]

\centering  
\subfigure{
\label{Fig3.sub.1}
\includegraphics[width=0.15\textwidth]{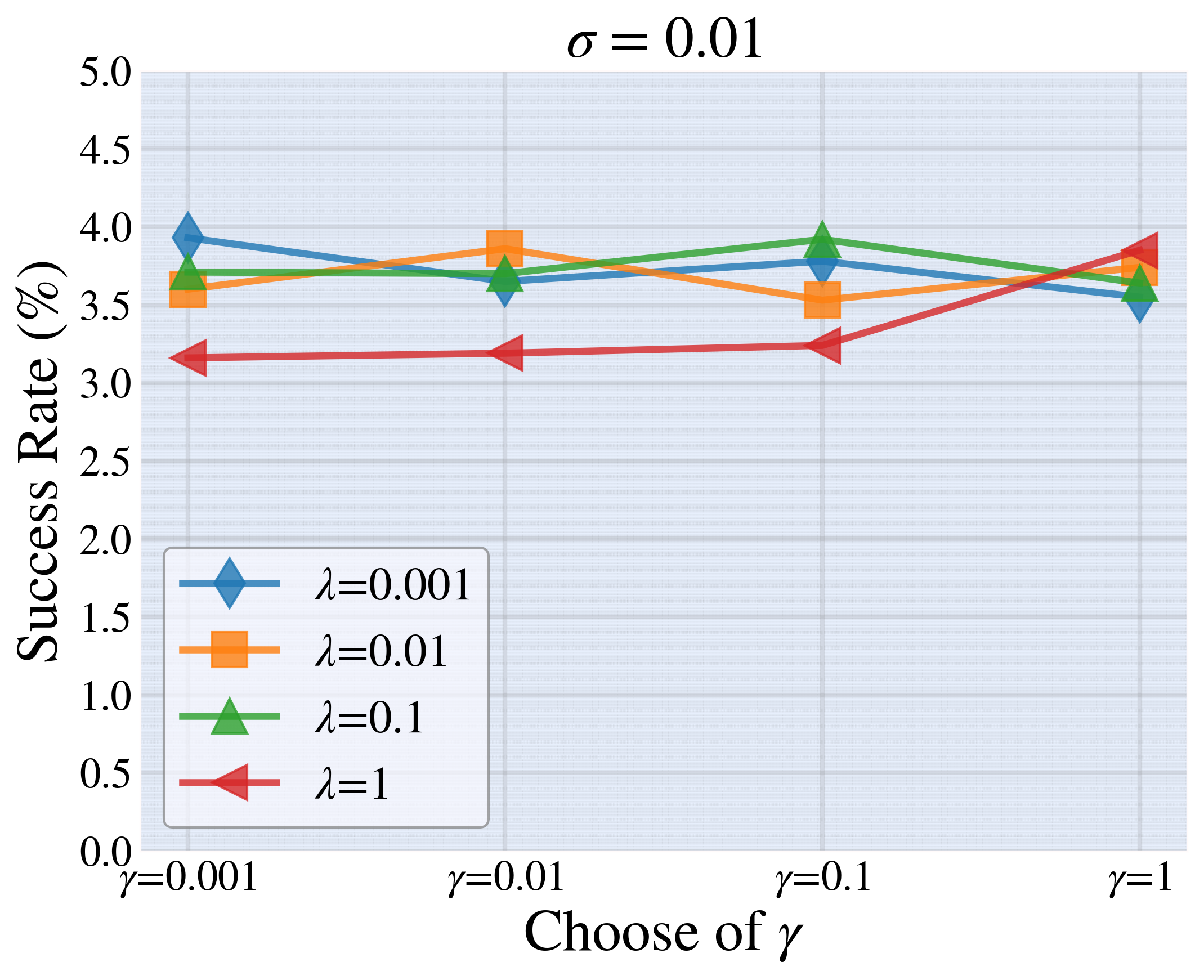}}
\subfigure{
\label{Fig3.sub.2}
\includegraphics[width=0.15\textwidth]{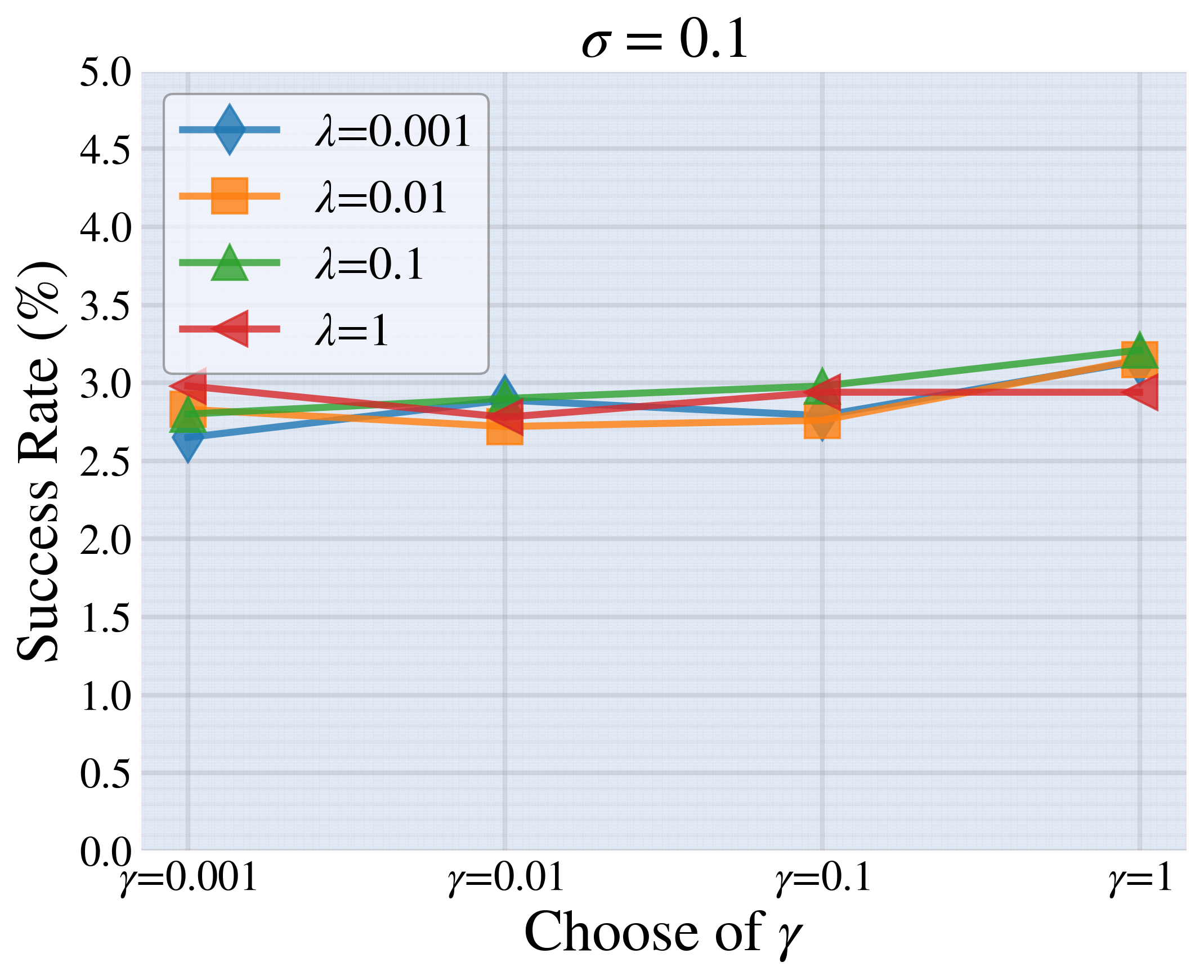}}
\subfigure{
\label{Fig3.sub.3}
\includegraphics[width=0.15\textwidth]{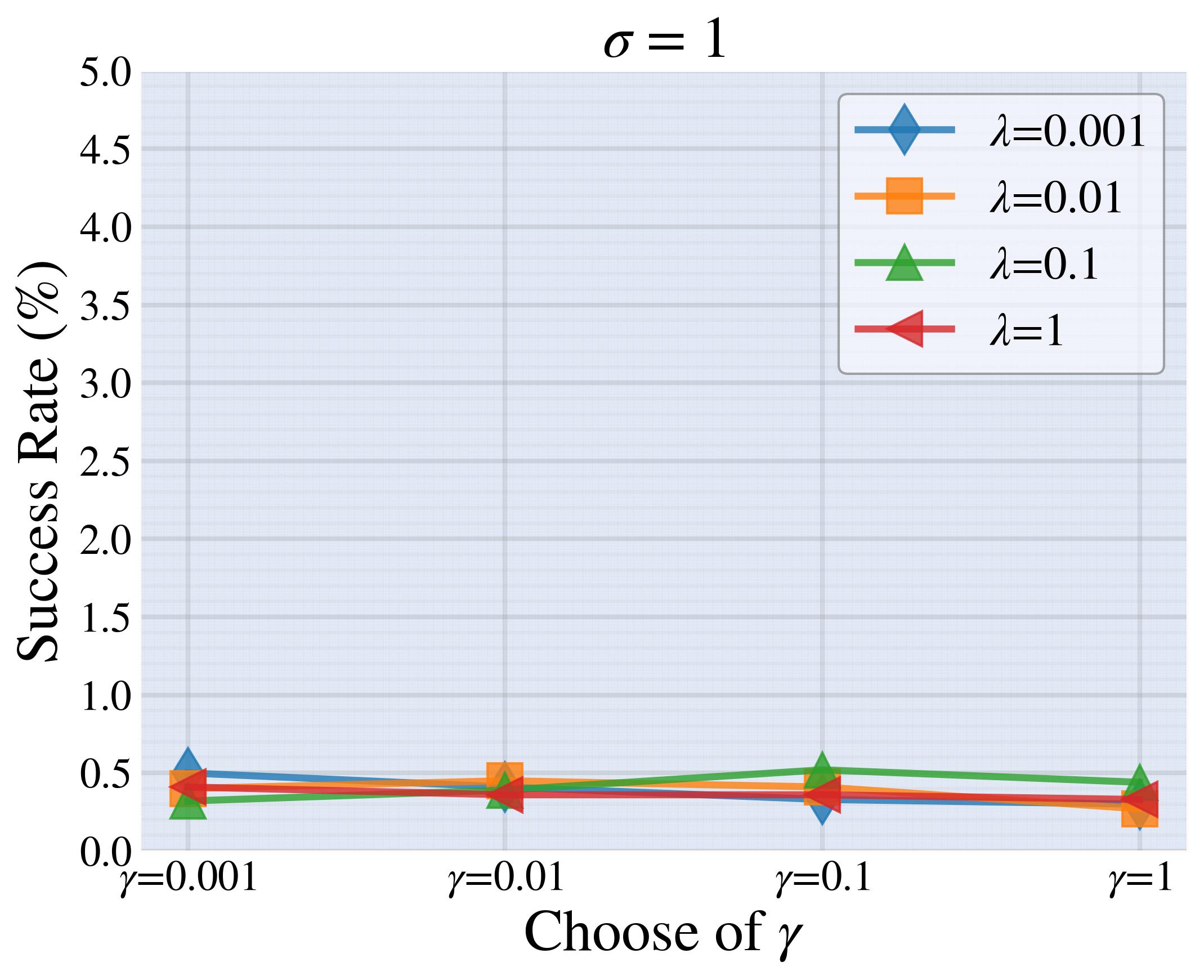}}
\vspace{-5pt}
\caption{Transfer success rates of MIM+IBTA with different parameter settings in the targeted scenario.}
\label{Fig 3}
\vspace{-5pt}
\end{figure}
\definecolor{mybrown}{RGB}{50, 80, 120}

\begin{table*}[h]
\vspace{-5pt}
  \centering
  \caption{Transfer Success Rate (\%) for transferable attacks, where the values in the table represent the results of 20/40/60 attack steps.}
  \tiny
  \setlength{\tabcolsep}{2.5pt}
  \renewcommand{\arraystretch}{0.8} 
    \begin{tabular}{c|ccccc||ccccc}
    \toprule[1pt]
    \multirow{2}[2]{*}{Attack} & \multicolumn{5}{c||}{\textbf{\textcolor{mybrown}{Src:Res50 (untargeted)}}} & \multicolumn{5}{c}{\textbf{\textcolor{mybrown}{Src:Dense121 (untargeted)}}}\\
     &{→Res50$^*$} & →VGG19bn & →Dense121 & →Res152  & AVG Transfer Rate  &{→Dense121$^*$} &  →VGG19bn & →Res50 & →Res152 &  AVG Transfer Rate\\
    
    \cmidrule{1-11}
   
    \rowcolor{gray!20}MIM  & {100.00/100.00/100.00} & 78.20/76.20/77.20 & 83.20/82.00/82.20 & 83.60/86.40/84.20 & 81.67/81.53/81.20 &  {100.00/100.00/100.00}   &  80.60/81.40/81.80 & 85.20/84.80/83.20 & 76.00/77.80/76.20 & 80.60/81.33/80.40 \\
      \rowcolor{blue!15} MIM+IBTA &  {100.00/100.00/100.00} & \textbf{80.40/82.40/81.20} & \textbf{86.60/85.20/86.80} & \textbf{86.80/89.60/89.60} & \textbf{84.60/85.73/85.80}  & {100.00/100.00/100.00}  & \textbf{84.00/84.80/84.40} & \textbf{87.00/87.20/86.80} & \textbf{79.40/80.40/81.80} & \textbf{83.47/84.13/84.30} \\
    \rowcolor{gray!20}TIM  & {100.00/100.00/100.00}   & 95.80/97.60/98.80 & 99.00/100.00/100.00 & 98.80/99.80/100.00 & 97.87/99.13/99.60 &  {100.00/100.00/100.00}   &  94.20/95.60/96.60 & 95.60/96.20/97.40 & 92.60/93.80/96.00 & 94.13/95.20/96.67  \\
    
    \rowcolor{blue!15}TIM+IBTA  & {100.00/100.00/100.00}   & \textbf{98.00/98.60/99.80} & \textbf{99.40/100.00/100.00} & \textbf{99.60}/99.80/100.00 & \textbf{99.00/99.33/99.93}  &  {100.00/100.00/100.00}   & \textbf{95.20}/95.60/\textbf{97.60} & \textbf{96.60/97.80/98.20} & \textbf{95.00/95.20/96.80} & \textbf{95.60/96.20/97.53} \\
    \rowcolor{gray!20}DIM & {100.00/100.00/100.00}  & 96.60/97.20/98.60 & 98.60/99.20/99.60 & 99.20/99.40/100.00 & 98.13/98.60/99.40   & {100.00/100.00/100.00}  & 95.60/96.60/96.00 & 96.60/97.60/98.20 & 93.40/95.40/96.60 & 95.20/96.53/96.93 \\
    \rowcolor{blue!15}DIM+IBTA  & {100.00/100.00/100.00}  & \textbf{97.40/98.60/99.20} & \textbf{99.00/99.60/99.80} & \textbf{99.40/100.00}/100.00 & \textbf{98.60/99.40/99.67}  & {100.00/100.00/100.00}  & 95.60/\textbf{97.00/97.20} & \textbf{97.00}/97.60/\textbf{98.80} & \textbf{94.40/95.80/97.40} & \textbf{95.67/96.80/97.80} \\

    \midrule[0.8pt]

    \multirow{2}[2]{*}{Attack} & \multicolumn{5}{c||}{\textbf{\textcolor{mybrown}{Src:VGG19bn (untargeted)}}} & \multicolumn{5}{c}{\textbf{\textcolor{mybrown}{Src:Res152 (untargeted)}}}\\
       &  {→VGG19bn$^*$}   & →Res50 & →Dense121 & →Res152 & AVG Transfer Rate   &  {→Res152$^*$}  & →VGG19bn & →Res50 & →Dense121 & AVG Transfer Rate\\
    
    \cmidrule{1-11}
   
    \rowcolor{gray!20}MIM   &  {100.00/100.00/100.00}  & 58.80/57.80/56.60 & 62.20/60.80/60.00 & 45.60/44.40/44.00 & 55.53/54.33/53.53   &   {100.00/100.00/100.00}   & 73.00/73.20/73.20 & 88.60/89.20/89.40 & 79.80/79.60/82.20 & 80.47/80.66/81.60 \\
    \rowcolor{blue!15}MIM+IBTA   & {100.00/100.00/100.00}  & \textbf{63.00/63.40/62.60} & \textbf{66.00/68.20/66.40} & \textbf{49.20/48.40/49.40} & \textbf{59.40/60.00/59.47}   & {100.00/100.00/100.00}   & \textbf{75.40/76.20/75.60} & \textbf{90.60/91.40/91.80} & \textbf{84.60/85.00/83.20} & \textbf{83.53/84.20/83.53} \\
    \rowcolor{gray!20}TIM    &  {100.00/100.00/100.00}    & 84.00/87.80/89.60 & 88.80/91.20/98.60 & 74.20/77.40/79.80 & 82.33/85.47/89.33    &  {100.00/100.00/100.00}   & 92.80/93.80/94.40 & 97.60/99.00/99.20 & 98.40/98.60/98.60 & 96.27/97.13/97.40 \\
    \rowcolor{blue!15}TIM+IBTA   & {100.00/100.00/100.00}   & \textbf{86.60/90.60/93.20} & \textbf{92.80/93.00}/98.60 & \textbf{79.20/83.40/84.60} & \textbf{86.20/89.00/92.13}   &  {100.00/100.00/100.00}    & \textbf{93.80/95.80/96.80} & \textbf{98.20/99.40/99.80} & 98.40/\textbf{99.20/99.20} & \textbf{96.80/98.13/98.60}  \\
    \rowcolor{gray!20}DIM   & {100.00/100.00/100.00}   & 81.80/84.40/86.20 & 84.80/87.80/90.80 & 67.00/70.80/74.60 & 77.87/81.00/83.87   & {100.00/100.00/100.00}   & 90.60/95.20/95.60 & 97.60/99.20/99.40 & 97.20/98.60/98.20 & 95.13/97.67/97.73 \\
    \rowcolor{blue!15}DIM+IBTA   & {100.00/100.00/100.00}   & \textbf{85.40/89.40/89.00} & \textbf{88.60/92.00/93.40} & \textbf{74.00/75.40/79.60} & \textbf{82.67/85.60/87.33}    & {100.00/100.00/100.00}  & \textbf{94.00/95.80/96.40} & \textbf{99.40/99.60/99.60} & \textbf{98.20/99.00/99.00} &  \textbf{97.20/98.13/98.33} \\

    \toprule[1pt]
    \multirow{2}[2]{*}{Attack} & \multicolumn{5}{c||}{\textbf{\textcolor{mybrown}{Src:Res50 (targeted)}}} & \multicolumn{5}{c}{\textbf{\textcolor{mybrown}{Src:Dense121 (targeted)}}}\\
       &  {→Res50$^*$}   & →VGG19bn & →Dense121 & →Res152  & AVG Transfer Rate   &  {→Dense121$^*$}  & →VGG19bn & →Res50 & →Res152 &  AVG Transfer Rate\\
    
    \cmidrule{1-11}
   
    \rowcolor{gray!20}MIM   & {100.00/100.00/100.00} & 1.56/0.96/0.73 & 3.29/2.38/2.29 & 3.58/3.24/3.00 &  2.81/2.19/2.01   & {100.00/100.00/100.00}  & 1.73/1.33/0.93 & 2.42/1.69/1.58 & 1.67/1.29/1.04 & 1.94/1.44/1.18 \\
    \rowcolor{blue!15} MIM+IBTA   & {100.00/100.00/100.00}  & \textbf{1.96/1.60/1.51} & \textbf{3.96/3.40/3.18} & \textbf{4.20/4.02/3.93} & \textbf{3.37/3.01/2.87}   &  {100.00/100.00/100.00}  & \textbf{2.53/1.98/1.58} & \textbf{3.38/2.87/2.76} & \textbf{1.87/1.98/1.71} & \textbf{2.59/2.28/2.02} \\
    \rowcolor{gray!20}TIM   &   {99.89/100.00/100.00}    & 13.24/18.36/19.02 & 26.78/38.64/43.91 & 25.11/36.33/42.13 &  21.71/31.10/35.02   &  {100.00/100.00/100.00}   & 8.64/10.87/10.53 & 13.91/16.87/19.62 & 8.76/11.62/12.98 &  10.44/13.12/14.38\\
    \rowcolor{blue!15}TIM+IBTA  &  {99.96/100.00/100.00}    & \textbf{14.51/21.71/23.09} & \textbf{28.91/45.71/48.76} & \textbf{26.51/43.13/47.73} & \textbf{23.31/36.85/39.86}   &  {100.00/100.00/100.00}   &  \textbf{9.84/12.60/13.22} & \textbf{15.29/20.69/23.64} & \textbf{10.78/14.51/16.47} &  \textbf{11.97/15.93/17.78} \\
    \rowcolor{gray!20}DIM   &  {99.96/100.00/100.00}  & 13.62/17.64/19.96 & 25.36/36.53/42.78 & 22.80/32.47/37.76 &  20.59/28.88/33.50   &  {100.00/100.00/100.00}   & 10.07/11.53/11.80 & 14.98/17.56/19.53 & 9.09/11.18/12.29 & 11.38/13.42/14.54 \\
    \rowcolor{blue!15}DIM+IBTA   &  {100.00/100.00/100.00}   & \textbf{14.24/22.62/23.76} & \textbf{25.89/42.71/46.31} & \textbf{23.76/39.51/45.09} &  \textbf{21.30/34.95/38.39}  & {100.00/100.00/100.00}    & \textbf{11.80/14.60/14.98} & \textbf{15.93/22.47/25.00} & \textbf{10.91/14.76/16.89} & \textbf{12.88/17.28/18.96} \\

    \midrule[0.8pt]

    \multirow{2}[2]{*}{Attack} & \multicolumn{5}{c||}{\textbf{\textcolor{mybrown}{Src:VGG19bn (targeted)}}} & \multicolumn{5}{c}{\textbf{\textcolor{mybrown}{Src:Res152 (targeted)}}}\\   &  {→VGG19bn$^*$}
      & →Res50 & →Dense121 & →Res152 & AVG Transfer Rate   & {→Res152$^*$} & →VGG19bn & →Res50 & →Dense121 & AVG Transfer Rate\\
    
    \cmidrule{1-11}
   
    \rowcolor{gray!20}MIM   & {100.00/100.00/100.00} & 0.60/0.40/0.40 & 0.62/0.29/0.29 & 0.33/0.20/0.20 & 0.52/0.30/0.30   & {100.00/100.00/100.00} & 1.16/0.93/1.02 & 5.93/5.07/5.00 & 3.36/2.42/2.36  & 3.48/2.81/2.79 \\
    \rowcolor{blue!15}MIM+IBTA   &  {100.00/100.00/100.00}  & \textbf{0.64/0.56/0.53} & \textbf{0.73/0.44/0.40} & \textbf{0.40/0.40/0.29} & \textbf{0.59/0.47/0.41}   &  {100.00/100.00/100.00}  & \textbf{1.60/1.56/1.04} & \textbf{7.22/7.73/6.89} & \textbf{4.38/3.53/3.38} & \textbf{4.40/4.27/3.77} \\
    \rowcolor{gray!20}TIM   &  {100.00/100.00/100.00}   & 3.67/2.73/2.89 & 4.51/3.62/3.67 & 2.02/1.44/1.53 & 3.40/2.60/2.70   &   {99.73/100.00/100.00}  & 10.09/12.44/13.22 & 27.20/39.13/44.18 & 21.73/29.73/33.89 & 19.67/27.10/30.43  \\
    \rowcolor{blue!15}TIM+IBTA   & {100.00/100.00/100.00}    & \textbf{5.29/4.44/5.07} & \textbf{6.87/6.27/6.18} & \textbf{2.44/2.53/2.31} & \textbf{4.87/4.41/4.52}   &  {99.91/100.00/100.00}   & \textbf{10.22/14.40/15.49} & \textbf{27.89/45.04/51.00} & \textbf{23.87/36.53/41.62} &  \textbf{20.66/31.99/36.04} \\
    \rowcolor{gray!20}DIM   & {99.91/100.00/100.00}   & 3.16/2.16/2.56 & 4.40/3.16/3.13 & 1.53/0.96/1.20 & 3.03/2.09/2.30   &  {99.58/100.00/100.00}  & 9.00/11.24/14.00 & 25.82/37.91/44.24 & 20.38/29.51/34.47 & 18.40/26.22/30.90 \\
    \rowcolor{blue!15}DIM+IBTA   & {100.00/100.00/100.00}   & \textbf{4.16/3.87/3.60} & \textbf{5.60/5.11/5.27} & \textbf{1.93/1.67/1.62} & \textbf{3.90/3.55/3.50}   & {99.87/100.00/100.00}   & \textbf{10.42/14.98/16.42} & \textbf{26.71/44.84/49.56} & \textbf{22.38/35.78/40.40} & \textbf{19.83/31.87/35.46} \\

    \bottomrule[1pt]
    \end{tabular}
  \label{tab1}

\end{table*}

\subsection{Analysis of Expandability}
In this section, we explore the expandability of IBTA as a general framework for seamless integration into existing attack methods. We selected MIM, DIM, and TIM as baselines. Following the same settings as in Section \MakeUppercase{\romannumeral3}-A, we conducted experiments using the randomly selected ImageNet subset. The images were first resized to $256\times 256$ and then center-cropped to $224\times 224$. The momentum factor $\mu$ for MIM was set to $0.1$, the probability $p$ for input diversity in DIM was set to $0.7$, and the kernel length for TIM was set to $7$. The experiments are under multiple step-size settings in both targeted and non-targeted scenarios. For targeted attacks, we used three different step sizes: $20$, $100$, and $300$. For non-targeted attacks, which are relatively easier compared to targeted attacks, we used three step sizes: $20$, $40$, and $60$. Following \cite{zhao2021success}, we use Res50, Res152, Dense121, and VGG19bn as source models sequentially and test the transfer success rate on the remaining three models. The results are presented in Table \uppercase\expandafter{\romannumeral1}.
Our method demonstrates improvement in both targeted and non-targeted settings, especially in the targeted setting where it achieves a maximum improvement of $6.07\%$. In the non-target setting, where DIM and TIM already perform well, our method still provides efficient improvement. Meanwhile, on the source model VGG19bn, where the vanilla methods are not so strong, our method still achieves an average improvement of over $4\%$. This confirms the effectiveness of our framework in transfer attacks and demonstrates the expandability of our framework with other existing methodes.

\begin{table}[t]
\vspace{-5pt}
  \centering
  \caption{Transfer Success Rate (\%) for non-targeted attacks.}
  \tiny
  \setlength{\tabcolsep}{2pt} 
  \renewcommand{\arraystretch}{0.6} 
    \begin{tabular}{c|ccccccc}
    \toprule[1pt]
    \multirow{2}[2]{*}{Attack} & \multicolumn{7}{c}{\textbf{\textcolor{mybrown}{Src:Inc-v3}}} \\
      &  {→Inc-v3$^*$}  & →Inc-v4 & →IncRes-v2 & →Adv-Inc-v3 & IncRes-v2ens & ViT & AVG \\
    
    \cmidrule{1-8}
   
    MIM  & {100.00} & 49.00 & 42.20 & 38.40 & 21.00 & {38.60} & 37.84 \\
    DIM & {100.00} & 70.20 & 63.80 & 45.00 & 25.80 & {39.20} & 48.80 \\
    SI-NIM & {99.80}  & 44.20 & 37.80 & 37.00 & 22.00 & {38.20} & 35.84 \\
    {ATTA} & {100.00} & {72.60} & {65.00} & {45.80} & {26.60} & {39.60} & 49.92 \\
    {VMI-CTM} & {99.60} & {79.40} & {67.20} & {45.60} & {26.00} & 
{39.60} & 51.56 \\
    {PAM} & {100.00} & {81.40} & {68.60} & {47.20} & {27.80} & {39.80} & 52.96 \\
    Logits & {100.00}  & 85.60 & 79.60 & 50.80 & 28.80 & {40.20} & 57.00 \\
    RAP+LS & {100.00} & 85.60 & 82.00 & 52.20 & 32.00 & {40.80} & 58.52 \\
    S²I-TI-DIM & {100.00} & 92.00 & 87.20 & 57.00 & 36.00 & {41.20} & 62.68 \\
    Logits+IBTA  & {100.00} & 88.60 & 80.00 & 51.40 & 31.20 & {42.40} & 58.72 \\
    RAP+LS+IBTA & {100.00} & 88.60 & 82.00 & 52.40 & 32.20 & {43.20} & 59.68 \\
    S$^2$I-TI-DIM+IBTA & {100.00} & 93.40 & 89.60 & 60.40 & 42.20 & {44.80} & 66.08 \\
    \midrule[0.8pt]
   \multirow{2}[2]{*}{Attack} & \multicolumn{6}{c}{\textbf{\textcolor{mybrown}{Src:Ensemble}}}\\
   &  {→Ensemble$^*$}  &  →Inc-v4 & →IncRes-v2 & →Adv-Inc-v3 & IncRes-v2ens & ViT & AVG\\

    \cmidrule{1-8}

    MIM & {99.20} & 70.40 & 61.80 & 43.80 & 26.40 & {40.20} & 48.52 \\
    DIM & {99.20} & 93.00 & 89.20 & 54.60 & 39.40 & {40.80} & 63.40 \\
    SI-NIM & {98.60} & 70.00 & 60.60 & 42.00 & 26.40 & {39.80} & 47.76 \\
    {ATTA} & {99.00} & {94.20} & {91.00} & {55.60} & {40.20} & {41.20} & 64.44 \\
    {VMI-CTM} & {99.40} & {96.80} & {91.40} & {56.20} & {41.20} & {41.40} & 65.40 \\
    {PAM} & {100.00} & {97.60} & {93.60} & {60.20} & {42.80} & {41.80} & 67.20 \\
    Logits & {100.00} & 98.00 & 96.20 & 63.00 & 43.80 & {42.00} & 68.60 \\
    RAP+LS & {100.00} & 99.20 & 98.60 & 63.60 & 46.40 & {42.40} & 70.04 \\
    S²I-TI-DIM & {99.80} & 99.80 & 99.60 & 73.60 & 60.40 & {42.60} & 75.20 \\
    Logits+IBTA & {100.00} & 99.20 & 97.60 & 64.40 & 48.20 & {43.60} & 70.60 \\
    RAP+LS+IBTA & {100.00} & 99.40 & 98.60 & 65.60 & 51.40 & {43.80} & 71.76 \\
    S²I-TI-DIM+IBTA & {100.00} & 99.40 & 99.60 & 78.80 & 65.40 & {44.80} & 77.60 \\
    \bottomrule[1pt]
    \end{tabular}
  \label{tab2}
\vspace{-10pt}
\end{table}

\begin{table}[t]
\vspace{-5pt}
  \centering
  \caption{Transfer Success Rate (\%) for targeted attacks.}
  \tiny
  \setlength{\tabcolsep}{2pt}
  \renewcommand{\arraystretch}{0.6} 
    \begin{tabular}{c|ccccccc}
    \toprule[1pt]
    \multirow{2}[2]{*}{Attack} & \multicolumn{6}{c}{\textbf{\textcolor{mybrown}{Src:Inc-v3}}} \\
      &  {→Inc-v3$^*$} & →Inc-v4 & →IncRes-v2 & →Adv-Inc-v3 & IncRes-v2ens & ViT & AVG \\
    
    \cmidrule{1-8}
   
    MIM & {99.32} & 0.40 & 0.36 & 0.16 & 0.09 & {0.21} & 0.24 \\
    DIM & {99.13} & 2.11 & 1.36 & 0.18 & 0.11 & {0.33} & 0.82 \\
    SI-NIM & {99.01} & 0.24 & 0.16 & 0.07 & 0.09 & {0.19} & 0.15 \\
    {ATTA} & {99.16} & {2.36} & {1.62} & {0.16} & {0.18} & {0.37} & 0.94 \\
    {VMI-CTM} & {99.23} & {2.82} & {1.58} & {0.20} & {0.13} & {0.32} & 1.01 \\
    {PAM} & {99.45} & {2.97} & {1.71} & {0.24} & {0.22} & {0.43} & 1.11 \\
    Logits & {99.81} & 3.56 & 1.96 & 0.16 & 0.16 & {0.46} & 1.26 \\
    RAP+LS & {100.00} & 4.16 & 3.16 & 0.20 & 0.13 & {0.51} & 1.63 \\
    S²I-TI-DIM & {99.67} & 10.33 & 9.11 & 0.84 & 0.47 & {0.73} & 4.30 \\
    Logits+IBTA & {100.00} & 3.69 & 2.80 & 0.27 & 0.20 & {0.61} & 1.51 \\
    RAP+LS+IBTA & {100.00} & 5.56 & 3.53 & 0.22 & 0.27 & {0.65} & 2.05 \\
    S²I-TI-DIM+IBTA & {100.00} & 11.13 & 10.22 & 0.91 & 0.56 & {0.82} & 4.73 \\
    \midrule[0.8pt]
   \multirow{2}[2]{*}{Attack} & \multicolumn{6}{c}{\textbf{\textcolor{mybrown}{Src:Ensemble}}}\\
    &  {→Ensemble$^*$} &  →Inc-v4 & →IncRes-v2 & →Adv-Inc-v3 & IncRes-v2ens & ViT & AVG \\

    \cmidrule{1-8}

    MIM & {98.57} & 1.78 & 1.11 & 0.16 & 0.13 & {0.26} & 0.69 \\
    DIM & {98.02} & 13.29 & 9.24 & 0.22 & 0.33 & {0.38} & 4.69 \\
    SI-NIM & {97.68} & 1.87 & 1.04 & 0.09 & 0.09 & {0.22} & 0.66 \\
    {ATTA} & {98.96} & {14.82} & {11.02} & {0.18} & {0.16} & {0.49} & 5.33 \\
    {VMI-CTM} & {98.82} & {18.32} & {16.68} & {0.39} & {0.47} & {0.45} & 7.26 \\
    {PAM} & {99.13} & {22.45} & {19.27} & {0.46} & {0.51} & {0.53} & 8.64 \\
    Logits & {99.78} & 37.84 & 26.71 & 0.67 & 0.67 & {0.73} & 13.32 \\
    RAP+LS &{100.00} & 46.07 & 31.27 & 0.62 & 0.60 & {0.77} & 15.87 \\
    S²I-TI-DIM & {99.01} & 41.22 & 36.56 & 2.96 & 3.00 & {1.85} & 17.12 \\
    Logits+IBTA & {99.82} & 40.87 & 30.42 & 0.73 & 0.80 & {0.98} & 14.76 \\
    RAP+LS+IBTA & {100.00} & 49.29 & 35.96 & 0.93 & 0.98 & {1.02} & 17.64 \\
    S²I-TI-DIM+IBTA & {99.26} & 42.67 & 37.98 & 3.28 & 3.36 & {2.25} & 17.91 \\
    \bottomrule[1pt]
    \end{tabular}
  \label{tab3}
\end{table}

\subsection{Comprehensive  Experiments}
Following \cite{qin2022boosting}, we conducted experiments on several additional models. We selected Inception-v3 (Inc-v3) \cite{szegedy2016rethinking} and an ensemble model composed of three models (Res152, Dense121, and Inc-v3) as the source models. As for the target models, we chose Inception-v4 (Inc-v4) \cite{szegedy2017inception}, Inception-ResNet-v2 (IncRes-v2) \cite{szegedy2017inception}, { Vision Transformer (ViT) \cite{DosovitskiyB0WZ21}}, as well as two adversarially trained models, Inc-v3$_{adv}$ and IncRes-v2$_{ens}$. 
In addition to the baseline methods mentioned earlier, we included several additional methods in the comparison, including SI-NIM \cite{lin2019nesterov}, Logit \cite{zhao2021success}, {ATTA \cite{WuSLK21}} as well as two state-of-the-art methods, {VMI-CTM \cite{Wang021}}, {PAM \cite{ZhangHWLWW0L23}}, RAP-LS \cite{qin2022boosting}, and S²IM \cite{long2022frequency}. For the newly added methods, we used the following parameter settings: for SI-NIM, the momentum factor $\mu$ was set to $1$, {iterations $K$, $K_{outer}$ and $K_{inner}$ for ATTA are all chosen as $10$, inner iteration number $N$ for VMI-CTM is set to $20$, while path number $n$ and copy number $m$ for PAM are set to $8$ and $4$ respectively.} for Logit, we combined it with TIM and followed the setting in \cite{zhao2021success}, choosing $T=300$ as its iteration steps, for RAP-LS, we followed the setting in \cite{qin2022boosting}, choosing $T=400$, $K_{LS}$=100 as the starting step of inner optimization, and setting the budget of perturbation for inner optimization $\epsilon_n=12$, for S$^2$IM, we combined it with DIM and TIM, and chose the tuning factor $\rho = 0.5$ according to \cite{long2022frequency}. Except for Logit and RAP-LS, we consistently set $T=20$ for all methods. {We applied the IBTA framework to the state-of-the-art methods Logit, RAP-LS, and S$^2$I-TI-DIM in order to validate that our framework can bring further improvements for both targeted and non-targeted attacks.}
The experimental results, presented in Tables \uppercase\expandafter{\romannumeral2} and \uppercase\expandafter{\romannumeral3}, demonstrate the efficacy of our framework in improving adversarial transferability in both target and non-target settings. Our framework yields an average improvement of over $2\%$ in the non-target setting and also enhances transfer performance in the target setting. Furthermore, our framework remains effective even when the source model is an ensemble model or the victim model is adversarially trained.

\section{Conclusion}
Building upon the IB concept, we derived a simple and efficient MILB and proposed a novel framework IBTA aimed at enhancing the transferability of adversarial attacks.
With the help of IB, IBTA fosters greater independence among adversarial perturbations across invariant features, thereby enhancing the transferability of adversarial attacks.
Experimental results demonstrated the effectiveness of IBTA when integrated with existing methods, leading to improved transferability performance for both non-targeted and targeted attack. 

\bibliographystyle{unsrt}
\bibliography{reference}

\end{document}